# Document AI: A Comparative Study of Transformer-Based, Graph-Based Models, and Convolutional Neural Networks For Document Layout Analysis

Sotirios Kastanas
University of Amsterdam
sam50249@gmail.com

Shaomu Tan
UvA, LTL Lab
s.tan@uva.nl

Yi He
Elsevier BV
y.he.1@elsevier.com

## ABSTRACT

Document AI aims to automatically analyze documents by leveraging natural language processing and computer vision techniques. One of the major tasks of Document AI is document layout analysis, which structures document pages by interpreting the content and spatial relationships of layout, image, and text. This task can be image-centric, wherein the aim is to identify and label various regions such as authors and paragraphs, or text-centric, where the focus is on classifying individual words in a document. Although there are increasingly sophisticated methods for improving layout analysis, doubts remain about the extent to which their findings can be generalized to a broader context. Specifically, prior work developed systems based on very different architectures, such as transformer-based, graph-based, and CNNs. However, no work has mentioned the effectiveness of these models in a comparative analysis. Moreover, while language-independent Document AI models capable of knowledge transfer have been developed, it remains to be investigated to what degree they can effectively transfer knowledge. In this study, we aim to fill these gaps by conducting a comparative evaluation of state-of-the-art models in document layout analysis and investigating the potential of cross-lingual layout analysis by utilizing machine translation techniques.

## 1 INTRODUCTION

The task of comprehending documents presents formidable challenges for both research and industry alike. Documents come in a multitude of formats, ranging from digitally born files to scanned versions. Examples of such documents include invoices, receipts, resumes, scientific articles, and others. Although the core information within these documents is typically conveyed through natural language, the diversity of formats, layouts, document quality, and complex template structures contribute to the complexity of document understanding.

Presently, companies employ labor-intensive and time-consuming processes to extract information from documents, incurring significant costs. These methods primarily rely on rule-based approaches [6], [11], [19] that demand extensive human effort or machine learning techniques [18], [23], [2], such as decision trees and support vector machines. However, these approaches yield unsatisfactory results, particularly when dealing with visually rich documents. Moreover, while large deep learning language models, like BERT [3] or RoBERTa [16], have been used for document understanding tasks, their performance is limited [30]. The main reason is that they rely solely on the textual information. However, this kind of documents, such as articles, forms, and business records encompass a wealth of additional information beyond the text alone. The layout and the image information of these documents play a crucial role in their interpretation.

Emerging as a novel research area, Document AI or Document Intelligence [1] aims to address these challenges by combining Natural Language Processing and Computer Vision techniques. This interdisciplinary area focuses on automating the analysis of various document types, including webpages, scanned-document pages, and digitally born files. This analysis involves automatic reading, understanding, classifying and extracting information from documents. In real-world scenarios it mainly includes several tasks like:

- **Document Layout Analysis:** This task is about structuring each document page by understanding the images, text, tables, and the positional relationships in the document layout.
- **Document Image Classification:** This task is about understanding document images and classifying them into different classes such as scientific articles, invoices, receipts, and others.
- **Document Visual Question Answering:** This task is about recognizing the textual part from digital-born documents (PDFs) or scanned document images, and answering questions in natural language by judging the internal logic of the recognized text.

This research endeavor centers on the investigation of document layout analysis. This task can be viewed from two distinct perspectives. The first perspective is as an image-centric task. It involves identifying the layout position of unstructured documents by generating bounding boxes and categorizing various elements within the document such as authors, paragraphs, tables, and titles. It is essentially an object detection problem [33], and various models with diverse architectures, including vision transformers [27], graph-based models [35], and object detectors with convolutional neural networks (CNNs) [13] have been successfully employed for this purpose. However, one major limitation of this approach is the potential overlap of bounding boxes. In the context of industrial applications, this limitation presents itself since object detectors with a high IoU value [22] and threshold are necessary for effectively structuring each document page and extracting accurate information, see Figure 4, Appendix A.

Alternatively, the second perspective of document layout analysis can be viewed as a text-centric task. In this approach, Document

AI capitalizes on the information derived from the document's image, layout, and text. It tackles the document layout analysis task by classifying individual words (token classification) or sequences of words (sequence classification). Taking inspiration from the transformers' architecture, Document AI introduces multimodal models (both transformer-based and graph-based) that combine the textual characteristics of tokens with the layout modality and the associated image features. This integration enables a comprehensive analysis of the document's content, leveraging both visual and textual cues. However, although promising, a significant limitation of this approach is the necessity for a language-specific tokenizer or language model for the text features, especially when dealing with multilingual content. To address this issue for transformer-based models, Document AI has introduced a multilingual transformer-based model, LayoutXLM [31], pretrained on 53 different languages and a unique language-independent transformer for document understanding, LiLT [28]. LILT has shown its potential to transfer knowledge by conducting the cross-lingual transfer: learn the task from English and subsequently apply that acquired knowledge to other languages. In addition, in the course of assessing LiLT's performance across various languages, a yet-to-be-answered question revolves around the prospect of leveraging machine translation to yield a more optimal solution by transmuting documents in the original language on which LiLT's training was predicated.

Several datasets have been developed for layout analysis in the field of Document AI, including DocBank [14], PubLayNet [34], DocLayNet [20], GROTOAP2 [26], and FUNSD [9]. These datasets are extensive collections of documents, which have been annotated either manually or through the use of annotation tools. They encompass various types of documents such as forms, magazines, tenders, and scientific articles, and encompass diverse elements within a document, such as author information, text, paragraphs, and abstracts. These datasets offer companies the opportunity to pretrain their models on large-scale data, enabling them to subsequently apply these models to their own document collections. Nevertheless, a comparative analysis of Document AI models with different architectures is currently lacking, and it remains unclear which specific model would be the most suitable for a company or research, based on the particular type of documents they possess.

This study presents a comprehensive comparison of state-of-the-art models with varying architectures in the document layout analysis task, examining both image-centric and text-centric approaches using benchmark layout analysis datasets, which is currently lacking. Additionally, we investigate the cross-lingual ability of a language-independent transformer by utilizing machine translation techniques. Specifically, we assess the performance of LiLT, by training it on English forms and evaluating its effectiveness on English and the translated forms from various languages, thereby investigating its cross-lingual capabilities. To the best of our knowledge, this is the first time that the option of machine translation models is utilized in the document layout analysis for mitigating the major limitation of multilingual content for document understanding. More specifically, we focus on finding the extent to which LiLT can transfer its learning in a zero-shot transfer learning scenario and whether machine translation offers a more promising approach for using the model. The code is publicly available at

https://github.com/samakos/Document-AI-

Our contributions are summarized as follows:

- We compare LayoutLMv3, Paragraph2Graph (as future work for GROTOAP2), and YOLOv5 using DocLayNet and GROTOAP2 datasets which encompass a variety of document types and we identify the most effective model for document layout analysis as image-centric task on similar document collections.
- We compare LayoutLMv3 and Paragraph2Graph using GROTOAP2 dataset and we identify the most effective model for document layout analysis as text-centric task on similar document collections.
- We explore the transferability of knowledge within English and multilingual forms from FUNSD and XFUND datasets through machine translation using LiLT in a zero-shot transfer learning scenario. To the best of our knowledge, this is the first time that machine translation is utilized for document understanding in the context of multilingual content.

## 2 RELATED WORK
### 2.1 Document Layout Analysis as Image-Centric Task

This approach primarily focuses on accurately determining the placement of various elements within documents, encompassing authors, paragraphs, keywords, abstracts, and other relevant components. It is predominantly an object detection task, where the main objective is to classify these element regions by providing bounding boxes and their associated categories. To address this challenge, a variety of models with diverse architectures have been employed, including object detectors with convolutional neural networks (CNNs), which have shown promising performance across several benchmark layout analysis datasets like DocBank, PubLayNet and DocLayNet. Recent advancements in document layout analysis have explored the utilization of vision transformers as a replacement for CNNs to achieve better results, as showcased in a study [8]. While vision transformers exhibit more promising results, it is important to consider the significant computational cost associated with their implementation. Another recent development in the field involves the application of graph neural networks [29], enabling effective capture and modeling of spatial layout patterns within documents. These research endeavors highlight the ongoing exploration of different techniques and models aimed at enhancing the understanding and analysis of document layouts. Despite the recent development of Document AI models with different architecture, a comparison between them is currently lacking and it remains unknown which is the most promising model for industry and research to be used in respect of different kind of documents.

Among the object detectors using CNNs fine-tuned on the DocLayNet dataset, M-RCNN [7], Faster-RCNN [21], and YOLOv5 [10] have demonstrated noteworthy performance, with YOLOv5 exhibiting the highest results. In a recent study [29], Paragraph2Graph, a novel graph-based model was introduced and compared against the aforementioned models, but YOLOv5 still achieves the highest mean Average Precision (mAP), a metric used for object detection tasks. Nevertheless, the authors did not endeavor to conduct a comparative analysis involving a vision transformer within the scope of their research. Notably, recent research [8] highlights the significant potential of LayoutLMv3 as a transformer-based model for layout analysis tasks. Motivated by these promising findings, this study aims to conduct a comprehensive investigation into the performance of LayoutLMv3 within the specific domain of layout analysis. Evaluation and comparison of LayoutLMv3 with the aforementioned models will be performed to assess its effectiveness and potential on the DocLayNet dataset.

Furthermore, it is important to acknowledge the existence of the GROTOAP2 dataset, which predominantly consists of scientific articles. However, there is currently a lack of research specifically focusing on object detection within this dataset. As part of this study, a comparative analysis will be conducted, comparing the performance of YOLOv5, LayoutLMv3, and Paragraph2Graph (as future work) on the GROTOAP2 dataset. This analysis will provide valuable insights into the respective capabilities and suitability of these models for object detection tasks within the dataset. There is a noticeable dearth of comparative analyses pertaining to Document AI models with different architectures, specifically in the context of benchmark datasets. In light of this research gap, we have deliberately chosen GROTOAP2 and DocLayNet as the datasets of interest in our study. These datasets have been selected due to their inclusion of a more extensive array of classes and different document categories, thus posing a greater challenge for layout analysis tasks.

## 2.2 Document Layout Analysis as Text-Centric Task

This approach revolves around leveraging all the modalities together, involving image, layout, and text information embedded within documents. In the domain of transformer based models, the advent of LayoutLM [30] and LayoutLMv2 [32] marked significant milestones in Document AI. These models aimed to mimic human comprehension of documents by incorporating both textual and visual features. Subsequently, LayoutLMv3 [8] emerged, expanding the visual backbone to encompass vision transformers. These multimodal models have surpassed traditional language models like BERT and RoBERTa, which were previously employed for document understanding.

In the realm of document layout analysis, graph-based approaches have shown great promise, particularly for industrial applications. These approaches treat each text box within a document as a node in a graph, with edges connecting text boxes belonging to the same layout region. Initially, in previous studies, Doc-GCN [17] and Doc2Graph [5] utilized graph neural networks (GNNs) to tackle the problem, employing word-label or entity-label classification. These models faced limitations due to their reliance on language-dependent tokenizers to handle the textual information. However, a recent breakthrough was achieved in [29], where a novel language-independent graph-based model was proposed. Notably, this model overcomes the need for specific language-tokenizers to extract text features as part of a node embedding, rendering it language-independent. This development holds significant implications, as it allows for a more flexible and adaptable approach to document analysis, especially for multilingual content. It has come to our attention that a comparative analysis between the state-of-the-art transformer-based and graph-based models, as discussed previously, in the context of document layout analysis as a text-centric task involving token classification is lacking in existing literature.

In this study, our objective is to compare the performance of Paragraph2Graph [29] as a graph-based model and LayoutLMv3 [8] as a multimodal transformer-based model on the GROTOAP2 dataset. By leveraging all available features, including text, image, and layout, we aim to gain insights into the strengths and limitations of each model within the context of document layout analysis.

## 2.3 Document AI for Multilingual Document Understanding

As mentioned, one major limitation of the transformer based models is that they are language-dependent, as they rely on language-specific tokenizers to handle textual features. Even for the graph-based models, [5], [17] rely on tokenizers that cover a specific number of languages. However, for graph-based models Paragraph2Graph [29] was introduced as the first language-independent graph-based model for document understanding and more specifically the document layout analysis task. To address this limitation for transformer-based models, the introduction of LayoutXLM [31] presented a solution. LayoutXLM, as a multilingual version of LayoutLMv2 [32], was pretrained on 53 different languages, allowing for language-agnostic document analysis. More recently, a language-independent transformer, LiLT [28], has been proposed. LiLT was pretrained on English-content and fine-tuned on various languages, demonstrating its adaptability to different languages, despite being originally trained on English content. However, the extent to which LiLT can transfer its acquired knowledge remains uncertain. The work conducted by [28] encompassed experiments involving zero-shot transfer learning and fine-tuning methodologies. They trained on English forms from FUNSD dataset and tested or fine-tuned on multilingual forms from XFUND dataset. In our view, machine translation can play crucial role for using the same language the model was trained. In this study, we examine whether machine translation offers a more promising approach for dealing with multilingual content and we try to find to what extent LiLT can transfer its knowledge by testing on English forms from FUNSD dataset and translated forms from XFUND dataset which encompasses documents in various languages. For the best of our knowledge this is the first time that machine translation models are utilized for the document layout analysis task and more generally in the field of Document AI.

## 3 METHODOLOGY

## 3.1 Models

*3.1.1 Transformer-Based model for both text-centric and image-centric tasks.*

- **LayoutLMv3:** We chose LayoutLMv3 for a multimodal transformer based model, as the optimal solution for document layout analysis, encompassing tasks that focus on both text-centric and image-centric aspects. Positioned as the latest advancement within the LayoutLM family of models, LayoutLMv3 represents the state-of-the-art approach capable of effectively addressing both of these tasks. It distinguishes itself from its predecessors, LayoutLM and LayoutLMv2, by eliminating the reliance on convolutional neural networks (CNNs) for image feature extraction, a crucial step in parameter reduction. Pretraining LayoutLMv3 is performed on the IIT-CDIP dataset [24], which comprises a collection of 11 million document images. To process inputs, LayoutLMv3 takes a document image along with its corresponding text and layout information, encoding this information into contextualized vector representations. To handle the textual component, LayoutLMv3 is initialized with the pretrained weights of RoBERTa. As for the image features, LayoutLMv3 employs an image tokenizer initialized from a pretrained image tokenizer in DiT [12]. Within the context of document layout analysis as a text-centric task, we approach the task involving token classification. By leveraging the layout, image, and text information with the aid of LayoutLMv3, we classify individual words into different classes such as author, paragraph, title, abstract, etc. On the other hand, for the image-centric task, we consider the problem as object detection, utilizing the visual capabilities of the DiT module within the LayoutLMv3 architecture. The model's architectuce is illustrated in Figure 5, appendix B.

*3.1.2 Graph-Based model for both text-centric and image-centric tasks.*

- **Paragraph2Graph:** We selected Paragraph2Graph as our preferred choice for document layout analysis task, encompassing both text-centric and image-centric objectives. This model, recently introduced, offers a language-independent graph-based approach specifically tailored for document layout analysis. The decision to adopt Paragraph2Graph was driven by its superior performance, as reported in the referenced study [29]. Comparative evaluations against other existing graph-based models, such as Doc-GCN [17] and Doc2Graph [5], demonstrated Paragraph2Graph's superior capabilities. A visual representation of the Paragraph2Graph architecture is illustrated in Figure 6, Appendix B.

*3.1.3 Object Detector with CNNs for image-centric task.*

- **YOLOv5:** We have chosen YOLOv5 as our object detector using convolutional neural networks (CNNs) due to its superior performance compared to Faster-RCNN and Mask-RCNN, as indicated by [2], [20] and [29]. CNNs are a type of artificial neural network specifically designed for tasks such as image segmentation and object detection. YOLO (You Only Look Once) is a state-of-the-art model for real-time object detection, and YOLOv5 is one of the prominent members within the YOLO model family. YOLOv5 encompasses five different models, namely YOLOv5s, YOLOv5m, YOLOv5l, YOLOv5n, and YOLOv5x. For our experiments, we have opted for YOLOv5s due to its smaller size and faster training capabilities. YOLOv5 was initially released in 2015 and has been pretrained on the COCO dataset [15]. This model takes images as input, divides them into regions, and then calculates probabilities and bounding boxes for each region. YOLOv5's architecture is illustrated in figure 7, Appendix B.

## 3.2 Datasets and Exploratory Data Analysis

*3.2.1 DocLayNet Dataset.* The DocLayNet dataset [20] is a recently introduced dataset in COCO format [15] annotated by humans. It distinguishes itself from other datasets such as PubLayNet [34], DocBank [14], and GROTOAP2 [26], which primarily consist of scientific articles, by encompassing more complex and diverse layouts from various domains. These domains include financial papers, tenders, laws, manuals, patents, and scientific articles.

Unlike other datasets, DocLayNet defines 11 distinct classes, namely caption, footnote, formula, list-item, page-footer, page-header, picture, section-header, table, text, and title. The distribution of the document categories is demonstrated in Figure 1.

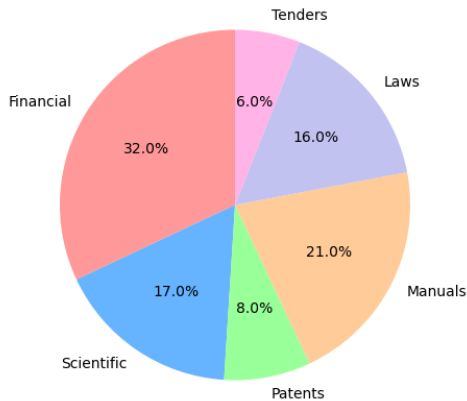

**Figure 1: Distribution of DocLayNet pages across document categories**

The majority of the pages in DocLayNet are in English (95%), but it also includes pages with multilingual content, such as German (2.5%), French (1%), Japanese (1%), and other languages. The DocLayNet dataset comprises a total of 80863 document pages, which have been split into three distinct subsets for different purposes. The training subset consists of 69103 document pages, while the validation subset contains 6480 pages, and the testing subset encompasses 4994 pages. This division of the dataset allows for effective training, evaluation, and testing of models on diverse document layouts and helps ensure robust performance across various scenarios. The distribution of each class within the documents is illustrated in Figure 2.

*3.2.2 GROTOAP2 Dataset.* The GROTOAP2 dataset consists of 13210 documents and 119334 document pages (PDF files), all of which are scientific articles. The ground-truth files for this dataset are in XML format. In comparison to the PubLayNet dataset, which is also comprised of scientific articles, GROTOAP2 was chosen due to its larger number of classes. PubLayNet includes five classes (text, title, figure, list, table), whereas GROTOAP2 contains 22 classes, namely abstract, acknowledgments, affiliation, author, bib-info, body-content, conflict-statement, copyright, correspondence, dates, editor, equation, figure, glossary, keywords, page-number, references, table, title, title-author, type, and unknown.

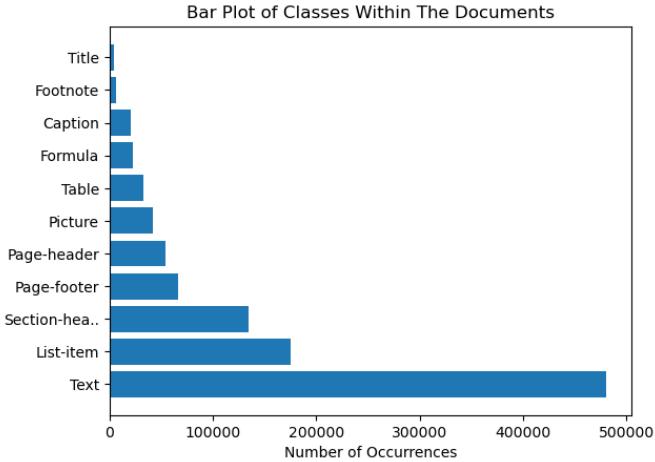

Figure 2: Number of instances for classes within the documents in DocLayNet dataset

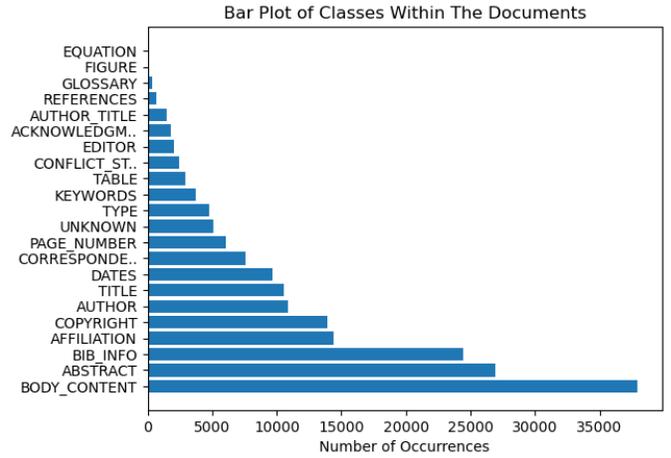

Figure 3: Number of instances for classes within the documents in the processed GROTOAP2 dataset

Although the DocBank dataset encompasses scientific articles and covers a wide range of classes (including abstract, author, caption, equation, figure, footer, list, paragraph, reference, section, table, and title), GROTOAP2 was preferred because we discovered that a portion of the DocBank dataset had incorrect bounding boxes, leading to misclassification of elements within a document.

The GROTOAP2 dataset was transformed from its original XML format to the COCO format to facilitate its utilization in document layout analysis as an image-centric task, specifically for object detection purposes. Additionally, we preprocessed the data to make it suitable for text-centric tasks involving token classification. In order to enhance the performance of models on specific classes that predominantly appear on the first pages of documents, we specifically utilized only the first pages of the document pages from the GROTOAP2 dataset and removed the others. The data preparation further involved splitting the PDF pages and converting them into document images for the models. This selective approach aimed to improve the models' accuracy and effectiveness in detecting and classifying those particular classes of interest. Following the selection of only the first pages from each document, the resulting dataset comprises 10492 document pages for training purposes. For validation, a subset of 1310 document pages was allocated, while an additional subset of 1317 document pages was set aside for testing. This refined dataset configuration allows for focused training, rigorous validation, and robust testing of models specifically on the initial pages of documents. The detailed steps for preparing the data to be used for both image-centric tasks (XML to COCO), text-centric tasks (XML to Token Classification), splitting the PDF pages and converting them into document images can be found at https://github.com/samakos/Document-AI-. The distribution of the different classes within the documents in the processed GROTOAP2 is illustrated in Figure 3.

*3.2.3 FUNSD and XFUND Datasets.* The FUNSD dataset encompasses a total of 149 training instances in the English language, complemented by an additional 50 instances designated for testing purposes. In the pursuit of probing the multifaceted linguistic competencies exhibited by various computational models, the research community introduced the XFUND dataset. This latter dataset extends its purview across multiple languages, namely French, Spanish, German, Italian, Japanese, Chinese, and Portuguese, thereby affording a broader cross-lingual spectrum of forms for analysis.

Both the FUNSD and XFUND datasets are presented in two distinct iterations: one configuration involves the delineation of 4 distinct classes (namely, "question," "answer," "header," and "other"), while the other configuration comprises 6 classes (namely, "B-question," "I-question," "B-answer," "I-answer," "header," and "other"). This present investigation, by deliberate design, centers its attention upon FUNSD and XFUND datasets within the confines of the 4-class schema.

## 3.3 Evaluation Metrics

*3.3.1 Evaluation Metrics for Layout Analysis as Image-Centric task.* To assess the performance of models in document layout analysis as an image-centric task (or object detection), we utilize the mean Average Precision (mAP) as our evaluation metric. This metric is derived from two fundamental measurements: precision and recall [25]. Precision and recall are defined as follows in equations (1), (2):

$$\text{Precision} = \frac{\text{TP}}{\text{TP} + \text{FP}} \quad (1)$$

$$\text{Recall} = \frac{\text{TP}}{\text{TP} + \text{FN}} \quad (2)$$

Here, TP, FP, and FN represent the counts of true positives, false positives, and false negatives, respectively. For object detection, we specifically consider the intersection over union (IoU) value, which measures the proximity of predicted bounding boxes to the actual bounding boxes. The IoU value can be expressed using the following mathematical equation (3):

$$\text{IoU} = \frac{\text{intersection}}{\text{union}} = \frac{Bg_t \cap Bp}{Bg_t \cup Bp} \quad (3)$$

where $Bg_t$ represents the ground truth bounding box and $Bp$ the predicted bounding box of the object detector.

By setting a manually defined IoU threshold, we determine TP, FP, and FN and subsequently calculate the precision. For our task, we compute the average precision (weighted mean of precisions at each threshold) for the 10 different IoU thresholds [0.5,0.95] with a step of 0.05 for classifying a detection as a true positive or not.

Finally, the mean average precision (mAP) (equation (4)) is the average of AP of each class.

$$\text{mAP} = \frac{1}{N} \sum_{i=1}^{N} AP_i \quad (4)$$

*3.3.2 Evaluation Metrics for Layout Analysis as Text-Centric task.* To assess the performance of models in document layout analysis as a text-centric task (or token classification in our usecase), we utilize the F1-score as our evaluation metric. In this task, instead of classifying regions and returning bounding boxes and each category, the objective is to assign specific labels to individual words, such as "text," "author," "paragraph," "abstract," and others. The F1-score incorporates the concepts of precision and recall, which are essential in assessing classification performance. Precision represents the proportion of correctly classified instances among all the instances predicted as positive for a particular label. Recall, on the other hand, signifies the proportion of correctly classified instances among all the actual positive instances for that label. By utilizing the definitions of precision and recall, the F1-score is calculated using the mathematical formula (5).

$$F1 = 2 \times \frac{\text{precision} \times \text{recall}}{\text{precision} + \text{recall}} \quad (5)$$

Finally, the F1-score as the harmonic average of the precision and recall is a suitable metric for imbalanced data, something that frequently occurs in the document layout analysis task.

## 3.4 Language-Independent Transformer & Machine Translation

LiLT (Language-independent Layout Transformer) is a novel transformer based model specifically designed for Document AI applications. Unlike other existing Document AI models, LiLT is capable of adapting knowledge obtained from English content and applying it to other languages. The architectural overview of LiLT can be observed in Figure 8, Appendix B. This model leverages both the textual and layout information extracted from a document image, incorporating these features into its framework to obtain enhanced representations. Various tokenization methodologies can be employed for the extraction of textual features, encompassing prominent options such as XLM-RoBERTa, InfoXLM, and others. Nevertheless, within the ambit of our investigation, the XLM-RoBERTa tokenizer was the selected modality. The language-independency of LiLT is achieved through the utilization of a Bi-directional attention complementation mechanism. The encoded textual and layout features are subsequently combined. In order to evaluate the adaptability of LiLT, we conducted fine-tuning for the document layout analysis task, treating it as a text-centric task involving token classification. LiLT was trained on the FUNSD dataset, concentrating on the four designated classes expounded upon in Section 3.2.2. Subsequently, the LiLT model underwent zero-shot evaluation across diverse languages encompassed within the XFUND dataset, despite lacking explicit training on them. Pertinent antecedents in the literature, notably referenced in [28], have indeed pursued analogous avenues of inquiry. It is, however, imperative to underscore the distinctiveness of our approach, wherein machine translation was harnessed as a pivotal instrumentality. The strategic utilization of machine translation was motivated by the objective of discerning whether it engenders a more efficacious trajectory, wherein documents, regardless of their native linguistic milieu, are seamlessly assimilated within an English-centric framework. To facilitate the translation process, we employed the M2M100 model [4], which encompasses 1.2 billion parameters and is available on the Hugging Face platform.

## 3.5 Implementation for Image-Centric and Text-Centric experiments

*3.5.1 Experiments for Document Layout Analysis as Text-Centric task.* To investigate document layout analysis as a text-centric task, our study employed the LayoutLMv3 and Paragraph2Graph models on the preprocessed GROTOAP2 dataset. We train on the training data as specified in 3.2 and we evaluate the models' performance on the validation set. For a fair comparison we trained the models until convergence. For Paragraph2Graph we adopted the hyperparameters specified in [29], which involved a learning rate of 0.001 and weight decay of 0.005. For LayoutLMv3 we adopted the hyperparameters specified in [8] involving a learning rate of 1e-5 which is considered the best for LayoutLMv3 in text-centric tasks.

*3.5.2 Experiments for Document Layout Analysis as Image-Centric task.* In order to investigate document layout analysis as an image-centric task, we conducted experiments utilizing the LayoutLMv3 model on the DocLayNet dataset. The aim of our study was to expand upon previous research conducted by [29], which did not include a transformer-based model. To ensure a fair comparison, we trained the model until convergence, following the methodology employed in the aforementioned study. During training, we utilized a batch size of 2 for 100 epochs. However, we implemented early stopping in cases where the model failed to exhibit any learning progress, thus terminating the training process. The learning rate chosen for our experiments was 2e-4, as proposed by [8] in their work on LayoutLMv3.

For the preprocessed GROTOAP2 dataset (as described in 3.2.2) we utilized YOLOv5 and LayoutLMv3 models. We plan to explore the use of Paragraph2Graph in future work. The models were trained until convergence, and subsequent evaluation was performed on the validation set. For LayoutLMv3 we use the same setting as for the experiment on DocLayNet. For YOLOv5 we chose YOLOv5s as a faster model. We use the same configuration files as the authors proposed without hyperparameter tuning. Although hyperparameter tuning is known to be effective in optimizing results, our study did not specifically focus on this aspect.

## 4 RESULTS

### 4.1 Document Layout Analysis as Image-Centric task

- **GROTOAP2:** The experimental results for document layout analysis, specifically focused on the image-centric task in the GROTOAP2 dataset, are presented in Table 1. As outlined, we employed LayoutLMv3 and YOLOv5 for the experiments, while leaving the exploration of Paragraph2Graph for future work. The performance comparison indicates that LayoutLMv3 exhibits superior performance in terms of overall mean average precision (0.751) when compared to YOLOv5 (0.725). Notably, LayoutLMv3 achieved convergence after approximately 7.5 epochs, while YOLOv5 required 75 epochs to reach convergence. Convergence was determined based on observations that further epochs did not yield decreases in the training and validation loss or increases in the mAP from validation data of the models. Noteworthy is the effective adaptation and capture of relevant information demonstrated by LayoutLMv3 within a relatively small number of epochs. However, it is important to consider the computational cost associated with LayoutLMv3, which necessitated 3 days to achieve convergence for 7.5 epochs. In contrast, YOLOv5 achieved convergence in only 2.5 hours for 75 epochs, utilizing a single GPU.

  From our perspective, YOLOv5 holds promise despite its lower performance and the requirement for a larger number of epochs to attain satisfactory results, primarily due to its speed advantages. Additionally, the analysis reveals that YOLOv5 outperforms LayoutLMv3 in 14 out of 22 classes. Specifically, the classes where YOLOv5 struggles in comparison to LayoutLMv3 include glossary, tables, figures, unknown elements, and page numbers. We believe that for these elements splitting the input image into visual tokens provide a better approach than using pixel arrays (CNNs). In all other cases, YOLOv5 demonstrates better or comparable performance. Finally, in terms of inference time, YOLOv5 proves to be a more favorable option, requiring less than 1 minute for the analysis of 1300 document images, whereas LayoutLMv3 necessitates approximately 7 minutes.

- **DocLayNet:** We present the results for document layout analysis as an image-centric task on DocLayNet dataset in Table 2. More specifically, we extend the research from [29] by adding the scores of LayoutLMv3 as a vision transformer. Based on [29], YOLOv5 needed approximately 65 epochs (approximately 1 day) to achieve convergence and note an overall score of 0.768 in terms of mAP. By contrast, Paragraph2Graph required only 5 epochs. We conduct experiment with LayoutLMv3 and as shown in Table 2 LayoutLMv3 achieves better performance than Paragraph2Graph but YOLOv5 remains the best. LayoutLMv3 needed only approximately 6 epochs to achieve the score using a single GPU but its experiment needed 3 days. Paragraph2Graph required approximately 2.5 days for the experiment. Paragraph2Graph achieves significant better results in specific classes like page-header, formula, page-footer, and section-header.

  In our view, all the models achieve approximately the same performance with YOLOv5 remaining in the first place. However, based on the time needed to be trained, even though YOLOv5 requires a very large number of epochs to converge, its advantage of training faster makes it better than the other models.

| mAP | YOLOv5 | LayoutLMv3 |
|---|---|---|
| BIB-Info | **0.884** | 0.845 |
| Body-Content | **0.893** | 0.853 |
| References | **0.918** | 0.834 |
| Affiliation | **0.907** | 0.878 |
| Page-Number | 0.481 | **0.713** |
| Abstract | **0.945** | 0.910 |
| Author | **0.854** | 0.837 |
| Dates | **0.866** | 0.802 |
| Title | **0.95** | 0.934 |
| Copyright | **0.934** | 0.881 |
| Acknowledgment | **0.874** | 0.858 |
| Unknown | 0.373 | **0.600** |
| Figure | 0.223 | **0.408** |
| Correspondence | **0.838** | 0.797 |
| Conflict-Statement | **0.775** | 0.738 |
| Table | 0.386 | **0.654** |
| Type | **0.809** | 0.808 |
| Keywords | 0.763 | **0.766** |
| Editor | **0.847** | 0.815 |
| Author-Title | 0.945 | **0.965** |
| Glossary | 0.491 | **0.631** |
| Equation | 0.000 | 0.000 |
| Total | 0.725 | **0.751** |

Table 1: Performance on GROTOAP2 for all categories

| mAP | YOLOv5 | LayoutLMv3 | Paragraph2Graph |
|---|---|---|---|
| Page-Header | 0.679 | 0.714 | **0.796** |
| Caption | 0.777 | **0.832** | 0.809 |
| Formula | 0.662 | 0.631 | **0.726** |
| Page-Footer | 0.611 | 0.629 | **0.920** |
| Section-Header | 0.746 | 0.709 | **0.824** |
| Footnote | **0.772** | 0.663 | 0.625 |
| Title | **0.827** | 0.769 | 0.643 |
| Text | **0.881** | 0.861 | 0.827 |
| List-Item | **0.862** | 0.834 | 0.805 |
| Picture | 0.771 | **0.799** | 0.581 |
| Table | **0.863** | 0.734 | 0.559 |
| Total | **0.768** | 0.743 | 0.738 |

Table 2: Performance on DocLayNet for all categories

## 4.2 Document Layout Analysis as Text-Centric task

The findings pertaining to document layout analysis as a text-centric task on the GROTOAP2 dataset are presented in Table 3. Comparative analysis of performance reveals that LayoutLMv3 outperforms Paragraph2Graph in terms of F1-scores. Specifically, LayoutLMv3 attains an F1-score of 0.866 with a mere 6 epochs dedicated to task adaptation. However, it is important to acknowledge the substantial computational burden associated with LayoutLMv3, necessitating approximately 2 days to complete the training process. In contrast, Paragraph2Graph yields an F1-score of 0.69, requiring 9 epochs for convergence, and demanding approximately 7 hours to complete the training process.

| F1-Scores | LayoutLMv3 | Paragraph2Graph |
|---|---|---|
| Total | **0.866** | 0.698 |

Table 3: Performance of LayoutLMv3 and Paragraph2Graph on GROTOAP2 dataset

## 4.3 Multilingual Document Understanding & Machine Translation

The findings of our study regarding the transfer learning capabilities of LiLT with the XLM-RoBERTa as tokenizer in a zero-shot transfer learning scenario and the results after translating the documents using the M2M100 translation model are presented in Table 4. Across all the conducted experiments involving diverse languages, it becomes evident that the efficacy of machine translation remains limited. Notably, the LiLT approach demonstrates superior performance when subjected to direct testing in distinct languages.

| LiLT | EN | DE | IT | ZH | JA | ES | FR | PT |
|---|---|---|---|---|---|---|---|---|
| Test | 0.48 | 0.49 | 0.38 | 0.35 | 0.37 | 0.39 | 0.50 | 0.44 |
| Machine Translation | - | 0.43 | 0.33 | 0.34 | 0.30 | 0.33 | 0.43 | 0.35 |

Table 4: LiLT performance on forms from XFUND dataset and the generated forms after machine translation

## 5 DISCUSSION

### 5.1 Document Layout Analysis as Image-Centric task

In our study, we conducted a comparative analysis of three distinct models, namely LayoutLMv3 (a vision transformer), YOLOv5 (an object detector with CNNs), and Paragraph2Graph (a graph-based model). We evaluated their performance on the DocLayNet dataset, while our investigation of Paragraph2Graph's efficacy on the GROTOAP2 dataset is planned for future work. Across both datasets, which encompass a wide range of document categories, including scientific articles in the case of GROTOAP2, our findings suggest that YOLOv5 exhibits the most promising potential for adoption in both research and industrial applications. Despite the notable requirement of numerous epochs to effectively adapt YOLOv5 to document analysis tasks and accurately detect various document elements, its exceptional speed and performance render the model capable of being trained using a single GPU within a day. Moreover, in terms of reference speed YOLOv5 outperforms all the other models with a significant difference. The backbone of using CNNs rather than vision transformers, exhibits significant difference in cost and speed and makes YOLOv5 more promising for industry and research to be used.

### 5.2 Document Layout Analysis as Text-Centric task

In our research study, we conducted a comparative analysis of two distinct models, namely LayoutLMv3, a multimodal transformer-based model, and Paragraph2Graph, a graph-based model. The purpose of our investigation was to evaluate their performance in document layout analysis, specifically focusing on token classification, within the preprocessed GROTOAP2 dataset, consisted of scientific articles. Our findings indicate that LayoutLMv3 showcases the most promising outcomes in the text-centric task of document layout analysis, particularly in relation to the token classification of first document pages in the GROTOAP2 dataset. Despite the considerable computational cost and convergence time associated with LayoutLMv3, the disparities in scores between the two models are substantial. To further advance our research, our future plan entails scrutinizing the specific elements within a document where Paragraph2Graph exhibits inferior performance compared to LayoutLMv3. However, it is important to note that LayoutLMv3's recent change in licensing to CC BY-NC-SA 4.0 renders the model unsuitable for industrial applications. Fortunately, an accessible alternative transformer-based option, LiLT, shows promise and we believe that similar results would be obtained.

### 5.3 Multilingual Document Understanding & Machine Translation

In our research, we explored whether using machine translation could be a better way to handle the document layout analysis task including token classification using a language-independent transformer, LiLT. We focused on translating documents using the same language that the LiLT model was trained on, in our case the English language. However, our results were unsatisfactory.

For all the different languages in the XFUND dataset, we noticed that the LiLT model worked better when we directly tested it on those languages, rather than using machine translation to turn the text into English first. This suggests that the quality of the machine translation plays a big role in how well this method works.

Our main focus was on translating individual tokens of the text. But translating each part into different languages creates problems because each language has its own unique rules for how words and sentences are put together. These challenges highlight the difficulties of using this approach. Exploring alternatives like Generative AI or different machine translators could be considered to determine if the particular machine translator is responsible for the unsatisfactory outcomes. Finally, we believe that treating the problem as a sequence classification problem rather than token classification would have provided more promising results.

## 6 CONCLUSION & FUTURE WORK

This study presents a comprehensive comparative analysis of recent sophisticated models employed for the document layout analysis task, including transformer-based models, graph-based models, and object detectors with CNNs. Our investigation establishes YOLOv5 as the most promising model for document layout analysis as image-centric task in terms of accuracy and speed, particularly for scientific articles as well as various document categories such as tenders, manuals, financial documents, laws, and patents. As a future direction, we intend to explore the performance of Paragraph2Graph in the GROTOAP2 dataset, although the computational cost associated with this model in respect to its performance on text-centric task should not be overlooked. Additionally, our research endeavors will involve the utilization of alternate versions of YOLOv5, as well as the most recent version, namely YOLOv8.

Furthermore, we have demonstrated that LayoutLMv3 exhibits great promise in layout analysis as a text-centric task specifically for scientific articles, outperforming the graph-based model, Paragraph2Graph. However, considering the recent licensing issue, we plan to investigate LiLT as a viable alternative for industrial applications. In our view, LiLT will provide similar results as LayoutLMv3.

Finally, we have introduced machine translation as a novel component in document layout analysis. Our results reveal that using machine translation didn't improve LiLT's performance when treating documents in the original language it was trained on. We think the quality of the machine translation is an important factor. As a next step, we plan to investigate whether switching to different machine translation models or addressing the problem as a sequence classification problem could lead to more promising outcomes.

# Appendix A  NECESSITY OF HIGH IOU VALUE AND THRESHOLD

Figure 4: The object detector classifies name Gregory Kucherov as an author with an IoU value of 93% without considering the letter v. The section and paragraph boxes exhibit overlapping regions. In both cases, the output is highly unsatisfactory, particularly for industrial applications.

## Appendix B  MODELS

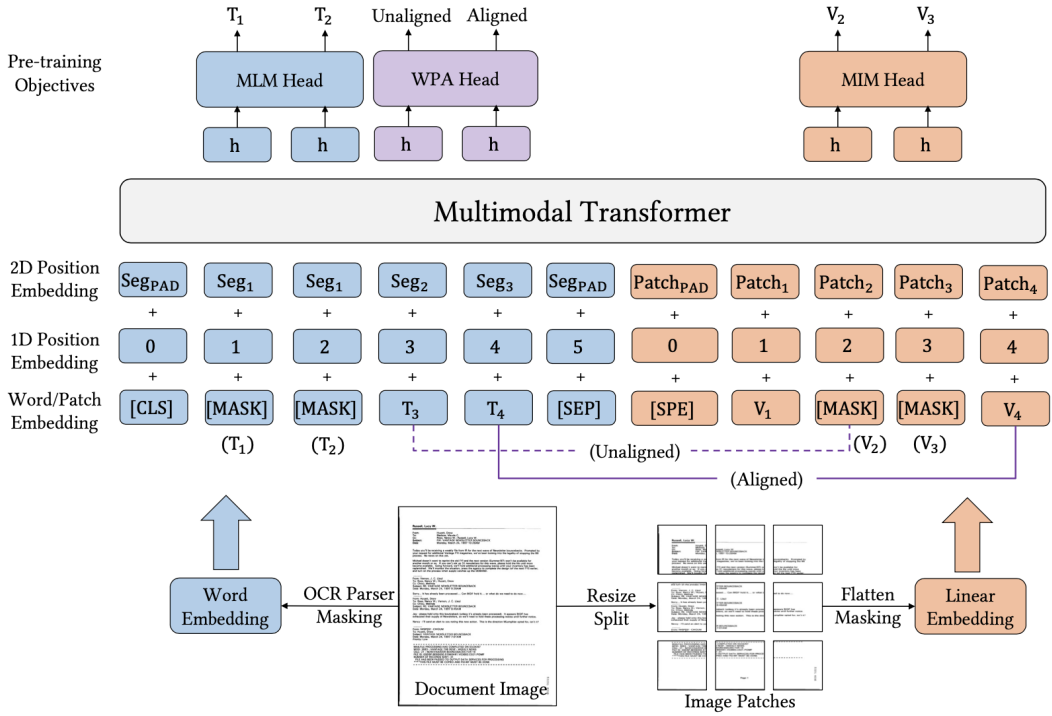

Figure 5: The architecture and pre-training objectives of LayoutLMv3. LayoutLMv3 is a pre-trained multimodal Transformer for Document AI with unified text and image masking objectives. Given an input document image and its corresponding text and layout position information, the model takes the linear projection of patches and word tokens as inputs and encodes them into contextualized vector representations. LayoutLMv3 is pre-trained with discrete token reconstructive objectives of Masked Language Modeling (MLM) and Masked Image Modeling (MIM). Additionally, LayoutLMv3 is pre-trained with a Word-Patch Alignment (WPA) objective to learn cross-modal alignment by predicting whether the corresponding image patch of a text word is masked. "Seg" denotes segment-level positions. "[CLS]", "[MASK]", "[SEP]" and "[SPE]" are special tokens.

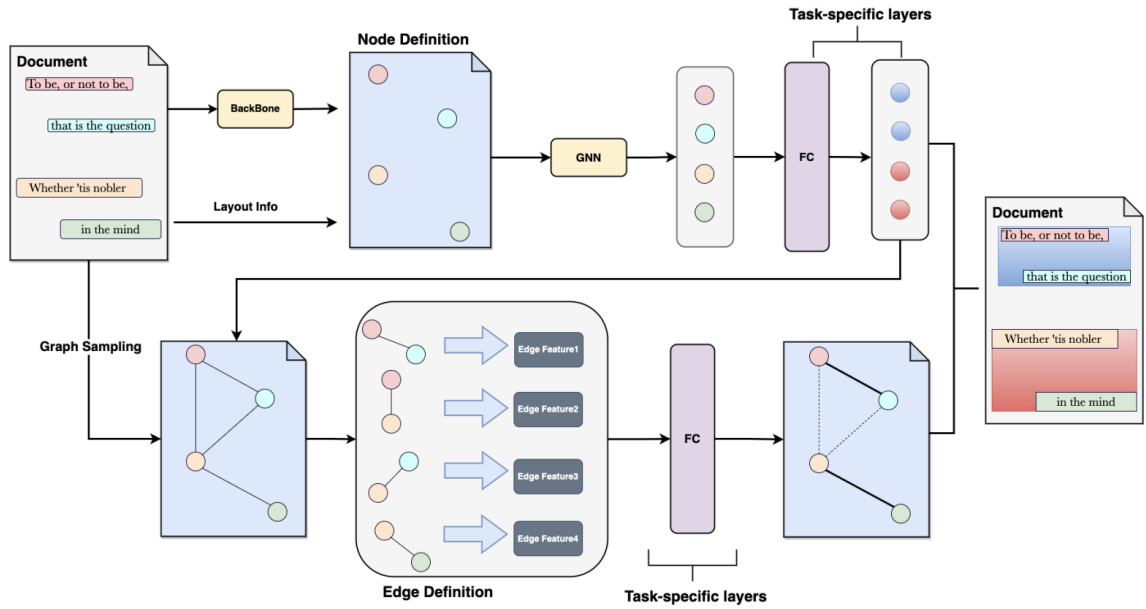

Figure 6: Paragraph2Graph architecture: The whole pipeline consists of five parts: node definition, edge definition, GNN, graph sampling and task-specific layers;dotted lines represent invalid edge connections

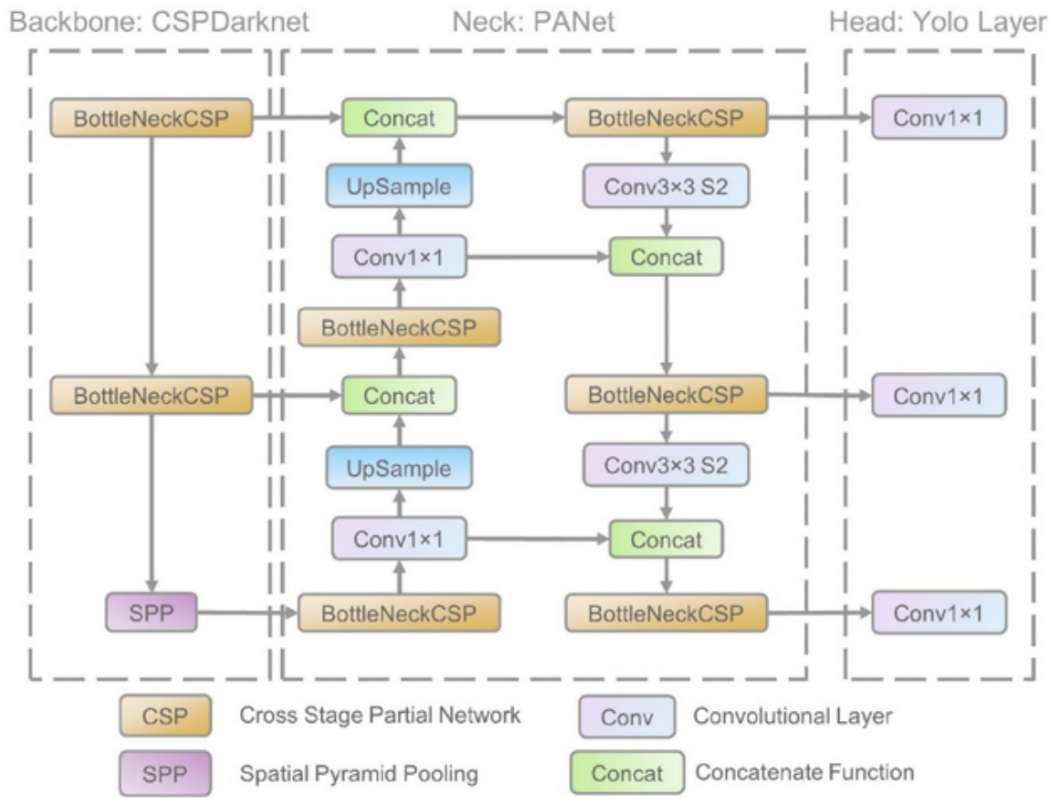

**Figure 7: YOLOv5 Architecture**

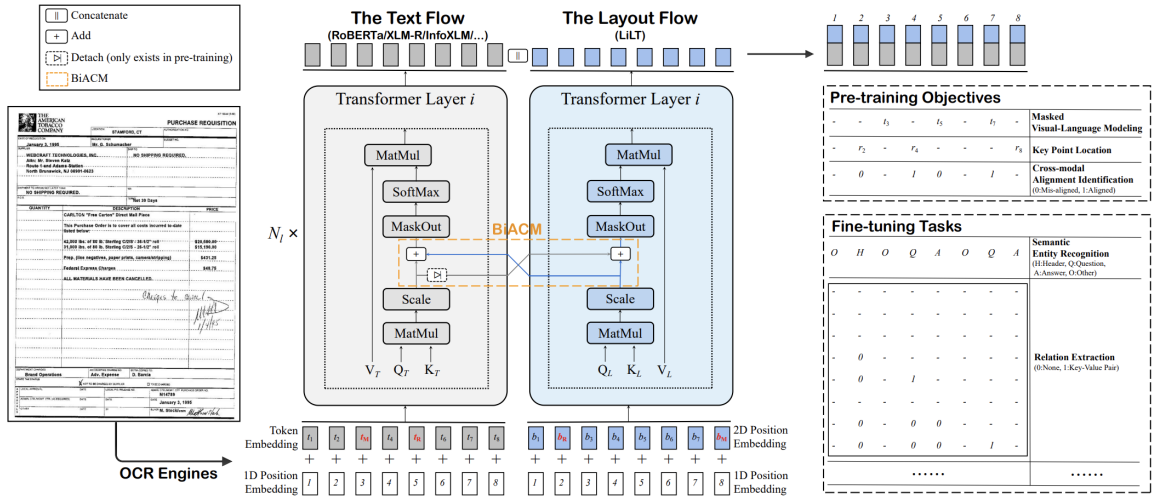

Figure 8: The overall illustration of LiLT framework. Text and layout information are separately embedded and fed into the corresponding flow. BiACM is proposed to accomplish the cross-modality interaction. At the model output, text and layout features are concatenated for the self-supervised pre-training or the downstream fine-tuning. Nl is the number of Transformer layers. The red *M/*R indicates the randomly masked/replaced item for pre-training. t, b and r represent token, box and region, respectively. Best viewed in zoomed-in.